\pdfoutput=1
\documentclass[11pt]{article}

\usepackage[utf8]{inputenc}
\usepackage[T1]{fontenc}
\usepackage{amsmath}
\usepackage{amssymb}
\usepackage{graphicx}
\usepackage{booktabs} 
\usepackage{hyperref} 
\usepackage{url}      
\usepackage{subcaption} 
\usepackage{color}
\usepackage[margin=1in]{geometry}
\usepackage{authblk}

\providecommand{\keywords}[1]{\textbf{\textit{Keywords:}} #1}

\begin{document}

\title{Towards a data-scale independent regulariser for robust sparse identification of non-linear dynamics}

\author[a]{Jay Raut}
\author[a,b,*]{Daniel N. Wilke}
\author[a]{Stephan Schmidt}

\affil[a]{\small Department of Mechanical and Aeronautical Engineering, University of Pretoria, Pretoria, South Africa}
\affil[b]{\small School of Mechanical, Industrial and Aeronautical Engineering, University of the Witwatersrand, Johannesburg, South Africa}
\affil[*]{Corresponding author: \texttt{daniel.wilke@wits.ac.za}}

\date{}
\maketitle

\begin{abstract}
Data normalisation, a common and often necessary preprocessing step in engineering and scientific applications, can severely distort the discovery of governing equations by magnitude-based sparse regression methods. This issue is particularly acute for the Sparse Identification of Nonlinear Dynamics (SINDy) framework, where the core assumption of sparsity is undermined by the interaction between data scaling and measurement noise. The resulting discovered models can be dense, uninterpretable, and physically incorrect. To address this critical vulnerability, we introduce the Sequential Thresholding of Coefficient of Variation (STCV), a novel, computationally efficient sparse regression algorithm that is inherently robust to data scaling. STCV replaces conventional magnitude-based thresholding with a dimensionless statistical metric, the Coefficient Presence (CP), which assesses the statistical validity and consistency of candidate terms in the model library. This shift from magnitude to statistical significance makes the discovery process invariant to arbitrary data scaling. Through comprehensive benchmarking on canonical dynamical systems and practical engineering problems, including a physical mass-spring-damper experiment, we demonstrate that STCV consistently and significantly outperforms standard Sequential Thresholding Least Squares (STLSQ) and Ensemble-SINDy (E-SINDy) on normalised, noisy datasets. The results show that STCV-based methods can successfully identify the correct, sparse physical laws even when other methods fail. By mitigating the distorting effects of normalisation, STCV makes sparse system identification a more reliable and automated tool for real-world applications, thereby enhancing model interpretability and trustworthiness.
\end{abstract}

\vspace{1em}

\section{Introduction}

The data-driven discovery of governing physical laws has emerged as a transformative paradigm in science and engineering, promising to automate the modeling of complex systems directly from measurement data. Among the various techniques developed for this purpose, the Sparse Identification of Nonlinear Dynamics (SINDy) framework has become a cornerstone method. SINDy's power lies in the core premise that the dynamics of many physical systems, when viewed in an appropriate coordinate system, are governed by only a few key interaction terms. This allows the governing Ordinary Differential Equations (ODEs) to be represented as a sparse linear combination of candidate functions from a predefined library. By leveraging sparse regression, SINDy produces parsimonious and interpretable models that naturally balance complexity and accuracy, thereby avoiding the challenge of overfitting, common to many black-box machine learning models~\cite{ref1}.

Despite its success, the practical application of SINDy is hindered by a fundamental reliance on magnitude-based sparse regression. The most common SINDy optimiser, Sequential Thresholding Least Squares (STLSQ)~\cite{ref1}, iteratively prunes terms from the candidate library based on the magnitude of their fitted coefficients. However, in numerous scientific and engineering contexts, measurements of different state variables can span vastly different scales. It is standard practice to normalise this data to a uniform range (e.g., [-1, 1]) to improve the numerical conditioning of the regression problem. While beneficial for numerical stability, this normalisation arbitrarily rescales the coefficients of the true underlying ODE. This distortion of the coefficient landscape directly undermines the logic of magnitude-based thresholding. The problem is further compounded by measurement noise, which can cause spurious, overfitted terms to acquire large magnitudes after normalisation, often exceeding those of the true, physically relevant terms.

This challenge has motivated research aimed at improving the robustness of the SINDy framework. These efforts can be broadly categorised into two main thrusts. The first focuses on robustifying the calculation of time derivatives from noisy data, which is a critical input to the SINDy regression problem. Significant research has been conducted on this front, often involving various data preprocessing~\cite {ref2}~\cite{ref5}. A prominent approach in this area is the weak-formulation of SINDy (WSINDy)~\cite{ref3}~\cite{ref7}~\cite{ref8}, which recasts the differential equation in an integral form, thereby avoiding direct numerical differentiation and providing orders-of-magnitude improvements in noise resilience, but sacrifices the computational benefits of using gradients. Another strategy involves using tools such as neural networks to learn a smooth, implicit representation of the data, from which clean derivatives can be obtained via automatic differentiation~\cite{ref4}~\cite{ref6}.

The second research thrust targets the sparse regression algorithm itself. This includes the development of more advanced optimisers, such as Sparse Relaxed Regularized Regression (SR3) and Stepwise Sparse Regression (SSR)~\cite{ref9} or Ensemble-SINDy (E-SINDy)~\cite{ref10}, which offer alternative pathways to a sparse solution but often still incorporate magnitude-based penalties in implementation~\cite{ref12}. Bayesian frameworks provide a more fundamental alternative, replacing deterministic thresholding with a probabilistic approach to term selection~\cite{ref14}~\cite{ref15}. By employing sparsity-promoting priors, such as spike-and-slab distributions, these methods can provide not only a sparse model but also a full posterior distribution for uncertainty quantification, albeit typically at a higher computational cost~\cite{ref13}~\cite{ref16}. Other recent work has explored group sparsity methods, which leverage structural similarities across multiple datasets to enhance identification, particularly for parametric systems~\cite{ref18}~\cite{ref20}~\cite{ref11}, and even more brute force methods ~\cite{ref17}.

While these advancements have significantly enhanced SINDy's robustness to noise, the specific, practical problem of sensitivity to data normalisation during the regression step remains largely unaddressed. Existing methods either tackle the inputs to the regression (derivatives) or replace the optimiser with more complex machinery. Still, they do not directly confront the failure mechanism induced by the interaction of normalisation, noise, and magnitude-based term selection. This paper introduces a direct, computationally efficient, and intuitive solution to this problem.

The primary contributions of this work are:
\begin{enumerate}
    \item A rigorous demonstration of how data normalisation fundamentally distorts the coefficient landscape in noisy SINDy problems, rendering magnitude-based thresholding unreliable.
    \item The proposal of the Sequential Thresholding of Coefficient of Variation (STCV), a novel magnitude-free sparse regression algorithm that bases term selection on statistical validity rather than absolute magnitude.
    \item Comprehensive benchmarking of STCV against established methods (STLSQ and E-SINDy), demonstrating its superior performance and reliability on normalised data across a suite of numerical simulations and a physical spring-mass-damper experiment.
\end{enumerate}

This work focuses on using SINDy and related methods to recover the correct model form rather than the exact parameters or coefficients, because (1) identifying the proper model form in itself improves the problem conditioning boosting the performance of any regression method used, and (2) the model form is still useful in applications such as anomaly detection (e.g., the model could be used to potentially detect and determine the source of anomalies).

This paper is structured as follows: Section \ref{sec:EvidenceOfDistortionDueToNormalisation} presents evidence that using SINDy with the STLSQ optimiser on normalised data distorts the estimated sparsity structure. To address this issue, Section \ref{sec:ProposedFramework} introduces the proposed STCV method, which uses a dimensionality property for sparsity identification. The performance benchmarks of STCV against established methods are provided in Section \ref{sec:Evidence}. The results and future outlook are presented in Section \ref{sec:DiscussionFutureOutlook} and the paper is concluded in Section \ref{sec:Conclusion}.

\section{The Distortion of Sparsity Structure by Data Normalisation}
\label{sec:EvidenceOfDistortionDueToNormalisation}

The standard SINDy framework seeks to identify a dynamical system of the form $\dot{\boldsymbol{x}} = f(\boldsymbol{x})$, where $\boldsymbol{x}(t) \in \mathbb{R}^{n \times 1}$ is the state of the system at time $t$, and $f: \mathbb{R}^{n \times 1} \mapsto \mathbb{R}^{n \times 1}$. Given time-series data of the state, $\boldsymbol{X} \in \mathbb{R}^{b \times n}$, and its derivatives, $\dot{\boldsymbol{X}} \in \mathbb{R}^{b \times n}$, where $b$ is the number of equidistant time points, the mapping is formulated as a linear system in the parameters:
\begin{equation} \label{eq:one}
    \dot{\boldsymbol{X}} = \boldsymbol{\Theta}(\boldsymbol{X})\boldsymbol{\Xi}
\end{equation}
Here, $\boldsymbol{\Theta}: \mathbb{R}^{b \times n} \mapsto \mathbb{R}^{b \times q}$ constructs a library matrix of $q$ candidate nonlinear functions over all $b$ time points. More specifically, each column of $\boldsymbol{\Theta}(\boldsymbol{X})$ is a candidate nonlinear function of the state variables (e.g., polynomials terms like $x_1, x_1^2, x_1 x_2$) for all $b$ time points, and $\boldsymbol{\Xi} \in \mathbb{R}^{q \times n}$ is a matrix of coefficients, with its element $\xi_{ij}$ indicating the contribution of the $j$th candidate term to the $i$th equation. The objective of SINDy is to produce a sparse coefficient matrix $\boldsymbol{\Xi}$ through the use of a sparse regression strategy.



The challenge arises when data normalisation is applied as a preprocessing step. Normalisation, defined here as scaling a signal so that its maximum absolute value is unitary, is a common practice to prevent numerical instability when state variables have disparate scales~\cite{ref22}. Note that this differs from the commonly used mean-centred normalisation, which would additionally distort the model form itself. To illustrate its effect, we consider the canonical Lorenz system:
\begin{equation}
    \label{eqn:Lorentz1}
    \begin{aligned}
    \dot{x} &= -10x + 10y \\
    \dot{y} &= 28x - y - xz \\
    \dot{z} &= -\frac{8}{3}z + xy
\end{aligned}
\end{equation}
In a noise-free scenario, normalizing the trajectory data $(\boldsymbol{X})$ is relatively benign. The process rescales the true coefficients in $\boldsymbol{\Xi}$ while perfectly preserving the correct sparsity structure, meaning the correct terms are identified, albeit with different numerical values.

The situation changes dramatically with the introduction of measurement noise. The process of normalisation, which depends on the specific data trajectory (e.g., the maximum value attained), acts as an implicit, uncontrolled hyperparameter that can catastrophically alter the outcome of the discovery process. This makes the algorithm's output dependent on an arbitrary preprocessing choice, fundamentally undermining the reliability and reproducibility of the scientific discovery.

To demonstrate this mechanism of failure, we add 0.1\% uniformly distributed noise to the Lorenz system trajectories and perform a non-sparse least-squares fit on both the original (unscaled) and normalised data using STLSQ. For the unscaled data, the mean values of the true coefficients remain large, whereas those of the spurious, overfitted coefficients are comparatively small. Their standard deviations are also low, making them easy to distinguish and prune with a magnitude-based threshold.

However, for the normalised data, the landscape is completely distorted. The mean magnitudes of the overfitted coefficients become comparable to, and in some cases larger than, the true coefficients. Furthermore, their standard deviations increase significantly. This means that, in any single realisation of noisy data, a spurious term is highly likely to have a larger coefficient than a true term, making it impossible for a magnitude-based method like STLSQ to correctly identify the model structure.

\begin{figure}[h!]
    \centering
    \begin{subfigure}{0.47\linewidth}
        \centering    
        \caption{}
         \includegraphics[width=0.8\linewidth]{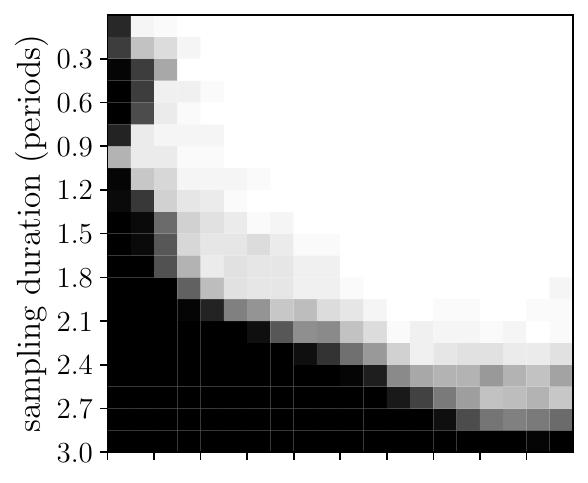}
    \end{subfigure}
    \begin{subfigure}{0.47\linewidth}
        \centering    
        \caption{}
        \includegraphics[width=0.8\linewidth]{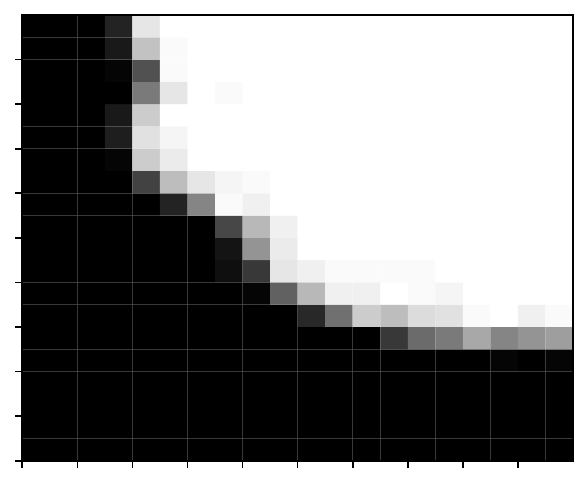}
    \end{subfigure}    
    \begin{subfigure}{0.47\linewidth}
        \centering
        \caption{}
         \includegraphics[width=0.8\linewidth]{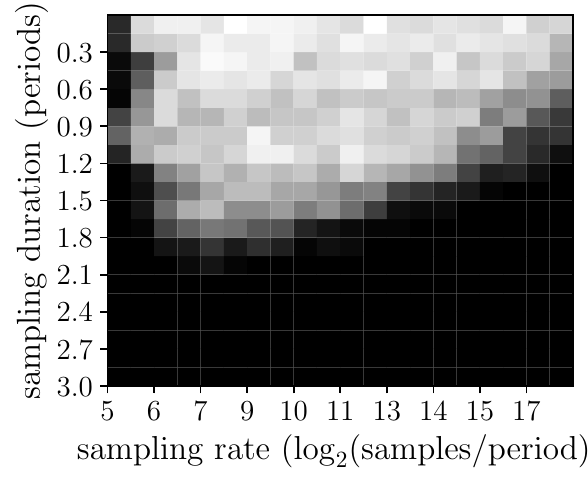}
    \end{subfigure}
    \begin{subfigure}{0.47\linewidth}
        \centering
        \caption{}
        \includegraphics[width=0.8\linewidth]{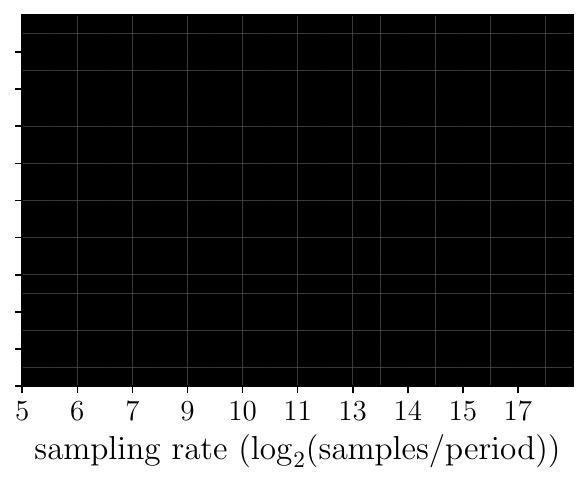}
    \end{subfigure}        
    \caption{Experiment demonstrating the degradation of SINDy sampling requirements for the Lorenz system in Equation \eqref{eqn:Lorentz1} when data is unscaled and normalised. The top and bottom rows respectively show noiseless data and data with 0.1\% uniformly distributed noise added. The left and right columns, respectively, show the results of SINDy on unnormalised and normalised data. White indicates a 100\% success rate in identifying the correct model sparsity, while black indicates a 0\% success rate. }
    \label{fig:lorenz}
\end{figure}

The practical consequence of this distortion is the complete failure of the discovery algorithm. Figure \ref{fig:lorenz} shows the success rate of STLSQ in identifying the correct Lorenz system equations as a function of data sampling rate and duration. For noiseless data, the success rate is lower when applying STLSQ to normalised data (top right) than to unscaled data (top left). This experiment is based on a benchmarking procedure found in Ref.~\cite{ref41}; refer to Appendix ~\ref{app:dataGen} for additional information on the data generation procedure. On unscaled noisy data (bottom left), STLSQ performs well across a wide range of sampling parameters. In stark contrast, on normalised noisy data (bottom right), STLSQ achieves a 0\% success rate across the entire tested parameter space. The algorithm terminates prematurely, retaining a dense set of incorrect terms because the spurious coefficients exceed the threshold and cannot be pruned. This provides incontrovertible evidence that a new, magnitude-free approach is required for robust system identification from normalised data.

\section{STCV: A Magnitude-Free Sparse Regression Framework}
\label{sec:ProposedFramework}
To overcome the limitations of magnitude-based thresholding, we propose a sparse regression framework that relies on a dimensionless statistical metric. The central hypothesis is that while the absolute magnitudes of coefficients are easily distorted by data scaling, their statistical consistency is a more robust indicator of physical relevance. Coefficients corresponding to true terms in the governing equations should be estimated consistently across different subsets of noisy data. In contrast, spurious coefficients, which are merely artifacts of noise, should vary erratically.

The notion of using a dimensionless, statistical parameter to label true underlying model terms is not novel, and has been previously explored in E-SINDy and UQ-SINDy~\cite{ref10}~\cite{ref13}. E-SINDy is a bootstrap aggregating (bagging) variant of STLSQ that significantly improves the sparsification accuracy. It achieves this by subsampling the data and/or library and fitting multiple sparse SINDy models. The inclusion probability across these models is used to label true terms. UQ-SINDy, on the other hand, performs Bayesian regression with specialised, non-Gaussian sparsity-inducing priors that help highlight the true underlying terms from the library, using the modal sharpness of the posterior distributions. UQ-SINDy employs the use of MCMC~\cite{ref19} to obtain posterior distributions of model coefficients. However, these methods possess limitations regarding robustness and computational efficiency. An alternative, robust statistical feature that is computationally efficient to evaluate is proposed.

\subsection{A Dimensionless Property for Sparsity Identification}

To quantify this statistical consistency, we employ the Coefficient of Variation (CV)~\cite{ref23}, a standard dimensionless measure of dispersion. For a given coefficient $\xi_{ij}$, the CV is defined as the ratio of its standard deviation $\sigma_{\xi_{ij}}$ to its mean $\mu_{\xi_{ij}}$, estimated over multiple model fits:
\begin{equation}
CV_{ij} = \frac{\sigma_{\xi_{ij}}}{\mu_{\xi_{ij}}}
\end{equation}
A low CV indicates high consistency (low relative variance), suggesting the term is physically significant. Conversely, a high CV indicates the term is an inconsistent artifact of noise. To create a metric that is directly proportional to a term's likelihood of being present and easier to interpret, we define the Coefficient Presence (CP) as a scaled reciprocal of the CV:
\begin{equation}
CP_{ij} = \frac{\sqrt{m} \cdot \mu_{\xi_{ij}}}{\sigma_{\xi_{ij}}}
\end{equation}
where $m$ is the number of data points used in the regression. This scaling makes the CP value independent of the sample size, ensuring that results are comparable across experiments with different data lengths. A high absolute CP value provides strong evidence that a term is present in the true underlying model. This trend is also observed in the UQ-SINDy study~\cite{ref13}, and we corroborated this in simple preliminary experiments.

The CP has a strong analogy to the inclusion-probability hyperparameter in the E-SINDy algorithm. E-SINDy’s inclusion probability is an MC method of evaluating the probability of a coefficient being non-zero by bootstrap aggregating the STLSQ method. On the other hand, the CP value indicates the likelihood that the coefficient is non-zero and is computed via closed-form BLR estimation with a Gaussian prior.

\subsection{The Sequential Thresholding of Coefficient of Variation (STCV) Algorithm}

The CP metric serves as the core of our proposed algorithm, Sequential Thresholding of Coefficient of Variation (STCV). STCV is an iterative refinement process that progressively sparsifies the candidate library based on the statistical validity of its terms. For computational efficiency, the required mean and standard deviation of each coefficient are not estimated via expensive Monte Carlo bootstrapping. Instead, we use Bayesian Linear Regression (BLR)~\cite{ref19} with a weak prior. This provides a closed-form analytical solution for the posterior mean and covariance of the coefficients, from which the CP values can be calculated directly and efficiently. 
The iterative process is detailed in Table \ref{tab:stcv_alg}.

\begin{table}[h!]
\caption{The Sequential Thresholding of Coefficient of Variation (STCV) Algorithm}
\label{tab:stcv_alg}
\centering
\begin{tabular}{p{0.9\textwidth}}
\toprule
\textbf{Algorithm 1: Sequential Thresholding of Coefficient of Variation (STCV)} \\
\midrule
\textbf{Inputs:} Design matrix $\boldsymbol{\Theta}$, derivative data $\dot{\boldsymbol{X}}$, STLSQ threshold $\lambda$, initial and final ridge penalties $\gamma_0, \gamma_f$, initial and final CP thresholds $\lambda_{CP,0}, \lambda_{CP,f}$, and number of steps $n_{steps}$. \\
\textbf{Output:} Sparse coefficient matrix $\boldsymbol{\Xi}$. \\

\textbf{Procedure:}
\begin{enumerate}
    \item Precondition data for numerical stability.
    \item Initialise:
    \begin{itemize}
        \item[a.] Perform an initial model fit using STLSQ with a near-zero threshold, $\lambda$, and initial ridge penalty, $\gamma_0$, to get an initial coefficient matrix, $\boldsymbol{\Xi}_{0}$.
        \item[b.] Use BLR to estimate the standard deviation $\sigma_{\xi_{ij}}$ for each coefficient in $\boldsymbol{\Xi}_{0}$.
        \item[c.] Calculate the CP value for each coefficient, $CP_{ij} = \mu_{\xi_{ij}}/\sigma_{\xi_{ij}}$.
        \item[d.] Apply an initial threshold: eliminate terms where $|CP_{ij}| < \lambda_{CP,0}$
        \item[a.] Create ramped schedules for the ridge penalty (from $\gamma_1$ to $\gamma_{n_{steps}}$ = $\gamma_f$) and the CP threshold (from $\lambda_{CP,1}$ to $\lambda_{CP,{n_{steps}}}$ = $\lambda_{CP,f}$).
    \end{itemize}
    \item Iterative Refinement:
    \begin{itemize}
        \item[b.] For each step $k$, from 1 to $n_{steps}$:
        \begin{itemize}
            \item[i.] Set current ridge penalty, $\gamma_k$, and CP threshold, $\lambda_{CP,k}$.
            \item[ii.] Inner Loop: While the model sparsity is changing:
            \begin{itemize}
                \item Fit a new model, $\boldsymbol{\Xi_{k}}$, using STLSQ on the currently active terms with ridge penalty $\gamma_k$.
                \item Estimate coefficient standard deviations using BLR.
                \item Calculate CP values for all active terms.
                \item Eliminate terms where $|CP_{ij}| < \lambda_{CP,k}$
                \item Check for convergence (i.e., if the set of active terms has changed).
            \end{itemize}
        \end{itemize}
    \end{itemize}
    \item Return $\boldsymbol{\Xi}=\boldsymbol{\Xi_{n_{steps}}}$.
\end{enumerate} \\
\bottomrule
\end{tabular}
\end{table}

While benchmarking the different methods, we noted that STCV sometimes fails by incorrectly retaining false terms and seldomly due to the incorrect elimination of true terms. Because the STCV and STLSQ methods are independent, STCV can be used as a pre-sparsification method for STLSQ, where STCV may partially sparsify the library, improving the problem's condition, and STLSQ can then complete the sparse regression to identify the final model form. This cascaded STCV-STLSQ could yield improved results and is additionally included in the performance benchmarking. The STCV-STLSQ method is performed as follows: The STCV is performed with a strong ridge penalty, and the resulting sparser library is used as an initial library for STLSQ. The strong ridge penalty makes the sparsification process of STCV very conservative, ensuring that a true term is not falsely eliminated. Subsequently, STLSQ is used to obtain the final terms. The precise code implementations of the methods are provided in Appendix~\ref{app:Code}.

\subsection{The Role of Regularization and Hyperparameter Selection}

In the STCV framework, the ridge penalty ($\gamma$) plays a nuanced role. It is not primarily used for regularization in the traditional sense of preventing overfitting by shrinking coefficients. Instead, its main purpose is to condition the least-squares problem, which stabilises the variance estimates that are critical for calculating reliable CP values. A large enough ridge penalty will overstabilise the problem such that many false terms possess CP values in comparable ranges to true terms.

The STCV algorithm employs a linear hyperparameter ramping strategy, akin to simulated annealing. This is necessary due to the hard-thresholding nature of the algorithm, where, upon eliminating a certain model term, it may no longer be recovered in subsequent iterations. It begins with a relatively high ridge penalty and a low CP threshold. This encourages a stable, conservative initial model. As the algorithm iterates and the library is refined, the ridge penalty is gradually decreased while the CP threshold is increased. This process gently guides the solution towards a sparse final model by parallelly reducing a stabilising prior while enforcing stricter statistical validity criteria at each step. Ideally, the ridge penalty should be ramped to zero since its role here is anti-sparsifying. However, in practice, a small ridge penalty may be necessary for numerical stability.

Like the thresholding hyperparameter of STLSQ and the inclusion probability threshold hyperparameter of E-SINDy, the CP thresholding hyperparameter of STCV is the main sparsity-promoting hyperparameter, such that a high CP threshold yields a sparser model, which requires manual tuning to produce models of the desired sparsity. In this study, we employ a grid search method to find the CP threshold that yields the correct model sparsity.

A distinct advantage of the CP threshold is that the value can be precisely tuned to control the exact number of active coefficients in the resulting model. Unlike the hyperparameters of STLSQ and E-SINDy, the CP threshold does not possess a finite logical range. The logical range of the STLSQ threshold is from 0 to the smallest coefficient of the underlying model, and the inclusion probability is from 0\% to 100\%. For STLSQ, this introduces a limitation: a false term cannot be eliminated if its coefficient exceeds the smallest true coefficient. In the case of E-SINDy, many terms may have an inclusion probability of exactly 0\% or 100\%, and partial elimination of this group is not possible.

\section{Performance Benchmarking and Analysis}
\label{sec:Evidence}
To validate the efficacy of STCV, we conducted a series of benchmarking tests comparing its performance with that of STLSQ and E-SINDy, with the data bagging option. The objective of this benchmark is to perform a hyperparameter grid search for each algortihm to identify whether there exists a set of hyperparameters that yields the correct underlying model form given measured dynamical data. The primary metric for comparison is the success rate, defined as the fraction of many trials (each with a unique noise realisation) in which an algorithm correctly identifies the exact sparsity structure of the underlying dynamical system. This is performed for increasing noise levels to evaluate the degradation of the various methods as a function of noise. Additionally, for each problem, the data are normalised to assess each algorithm's sensitivity to data scaling.

The general benchmarking procedure is as follows. Data are generated for each problem by numerically integrating the governing ODE over a specified time interval. The variable trajectories of the simulated system are injected with multiple instances and levels of noise, where the noise level is a percentage value of the recorded maximum absolute value of the variable trajectory. This produces multiple noisy datasets at varying noise levels. To produce a normalised dataset, the individual trajectories of each dataset are scaled such that the maximum absolute value is unity. The exact details of the data generation procedure for each problem (ODE, solver, simulation sampling rate and duration, etc., used) are listed in Appendix ~\ref{app:dataGen}.

For each noisy data instance, each algorithm is used to perform sparse regression with varying hyperparameters, and any instance that correctly identifies the underlying model form is recorded as a success, corresponding to that data. For each noise level, this modelling success is evaluated on each noisy data instance to produce a success rate. This benchmark presents the success rate of each method as noise levels increase. Each method is given the best chance of success by analysing the widest possible hyperparameter range. The exact algorithmic setup and range and distribution of hyperparameters used for each algorithm, for each problem, are listed in Appendix ~\ref{app:modelGen}.

\subsection{Canonical Dynamical Systems}

\begin{figure}[h!]
    \centering
    \begin{subfigure}[b]{0.49\textwidth}
        \centering
        \caption{Lorenz}        
        \includegraphics[width=\textwidth]{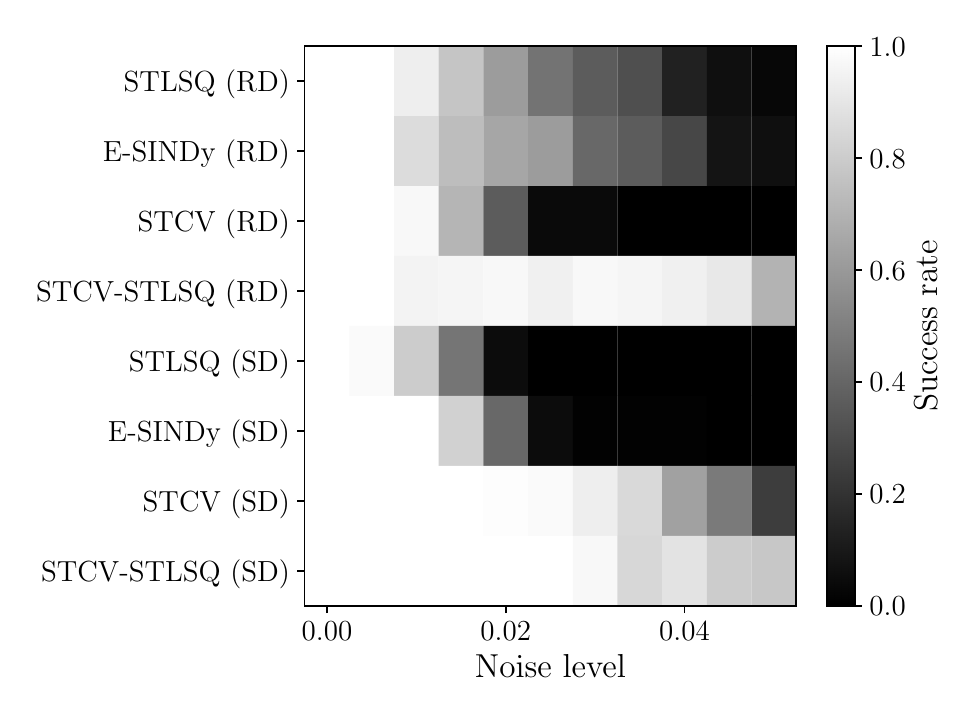} 
        \label{fig:fig4_lorenz}
    \end{subfigure}
    \hfill
    \begin{subfigure}[b]{0.49\textwidth}
        \centering
        \caption{Rössler}        
        \includegraphics[width=\textwidth]{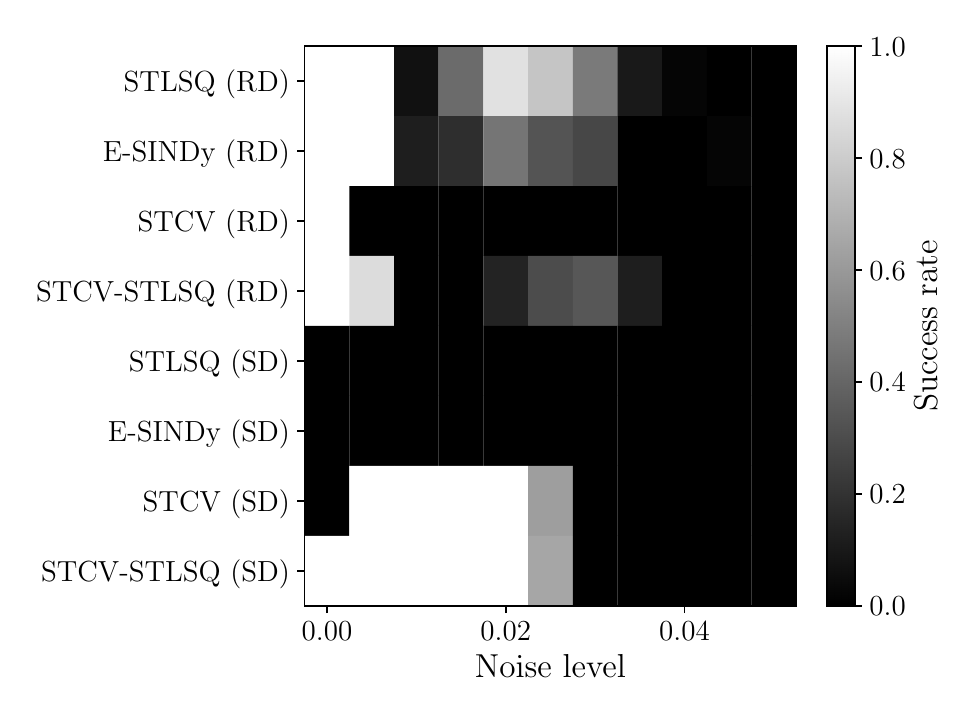}
        \label{fig:fig5_rossler}
    \end{subfigure}
    \begin{subfigure}[b]{0.49\textwidth}
        \centering
        \caption{Van der Pol}        
        \includegraphics[width=\textwidth]{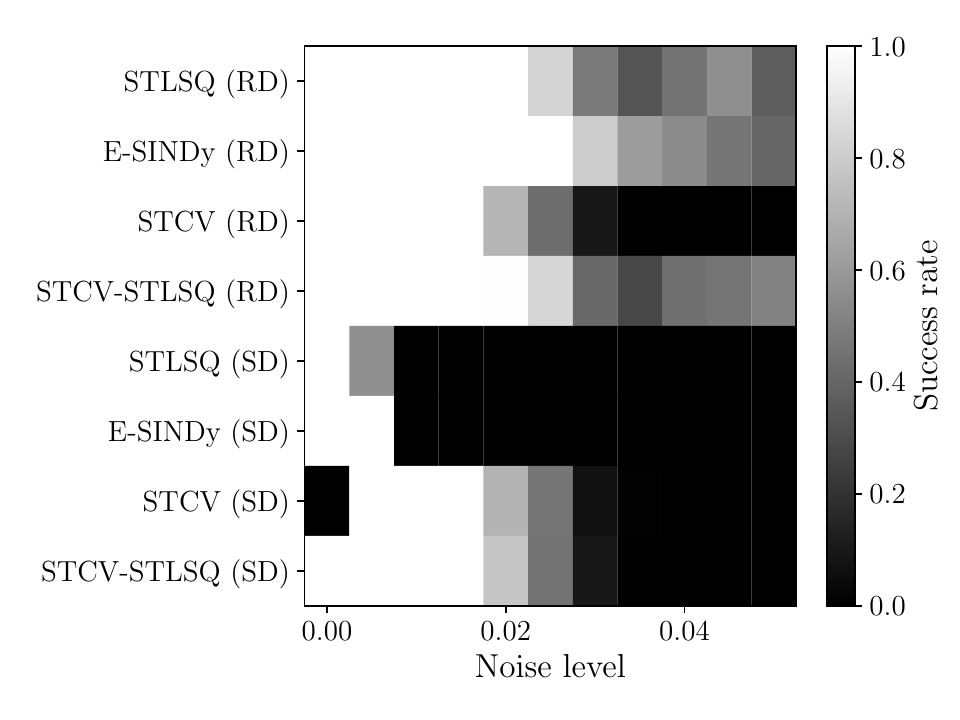}
        \label{fig:fig6_vdp}
    \end{subfigure}
    \hfill
    \begin{subfigure}[b]{0.49\textwidth}
        \centering
        \caption{Duffing}        
        \includegraphics[width=\textwidth]{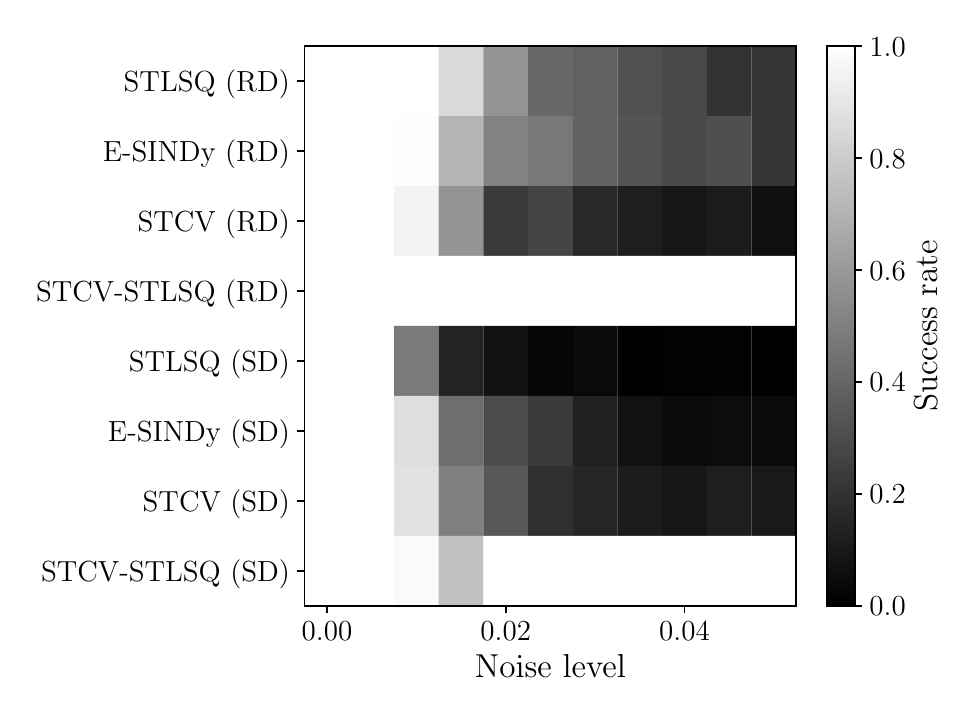}
        \label{fig:fig7_duffing}
    \end{subfigure}    
    \caption{Success rate results from the Lorenz, Rössler, Van der Pol, and Duffing oscillator models on the Raw Data (RD) and the Scaled Data (SD). The model form and the parameters used may be found in Appendix ~\ref{app:dataGen}. }
\end{figure}

We first tested the algorithms on four canonical systems commonly used in the SINDy literature: the Lorenz, Rössler, Van der Pol, and Duffing oscillators (Figures \ref{fig:fig4_lorenz}, \ref{fig:fig5_rossler}, \ref{fig:fig6_vdp}, and \ref{fig:fig7_duffing}). The results from our investigations consistently demonstrated a clear pattern. For unscaled (original) data, the performance of STCV was generally comparable to that of STLSQ and E-SINDy. However, when the trajectory data were normalised, the performance of STLSQ and E-SINDy deteriorated dramatically as noise levels increased. In contrast, STCV's performance remained significantly more robust, often maintaining a high success rate in noise regimes where the other methods failed. This directly validates our central hypothesis that a magnitude-free approach improves performance for normalised datasets.

A key observation is the shared point of absolute failure for the STLSQ and E-SINDy methods. This is due to the internal use of STLSQ in E-SINDy and highlights how bootstrap aggregating improves upon STLSQ while inheriting a fundamental limitation of STLSQ.

One observed anomaly was the occasional failure of STCV on noise-free normalised data for some systems. This is an artifact of the numerical integration schemes used, which can inject dynamically consistent numerical errors. Without the randomizing effect of noise, these spurious terms can overfit to these numerical errors with high consistency (i.e., have very low variance), making them indistinguishable from true terms for a purely statistical method. The addition of some noise is seemingly sufficient to break these near-perfect correlations and resolve the issue.

\subsection{Application to Engineering Systems: Where Normalisation is a Necessity}

While canonical systems provide a useful baseline, the true test of STCV lies in practical engineering problems where data normalisation is not merely an option but a necessity for numerical stability.

\subsubsection{Damaged Bearing Simulation}
To test the algorithms on a system where normalisation is a numerical necessity, we set up a damped mass-spring-damper system with forcing, designed to mimic the dynamics of a damaged bearing housing. The governing equations for the system are given as

\begin{equation}
    \label{eqn:bearing}
\begin{aligned}
    \dot{s}(t) &= v(t) \\
    \dot{v}(t) &= -10^6 s(t) - 2 \cdot 10^3 v(t) + F(t)
\end{aligned}
\end{equation}

where $s$ and $v$ represent the system displacement and velocity, respectively, and $F(t)$ is a step function of initial amplitude 0.5 N/kg and width 200 µs, followed by an amplitude of 0 N/kg thereafter. This forcing accelerates the system to simulate a rolling element passing over a defect, which momentarily changes the load on the bearing housing.

The parameters in Equation (10) have been tuned to match the response recorded on the housing of bearing 1 from the IMS dataset~\cite{ref33}, ensuring the model provides a realistic response. However, the exact forcing step size does not matter for this investigation as the data is necessarily normalised.

This high-stiffness system (note the $10^6$ coefficient for $s$) yields displacement signals that are orders of magnitude smaller than the velocity signals (a scale difference of approximately 30,000 times), rendering the unscaled regression problem numerically ill-posed and necessitating normalisation.

\begin{figure}[h!]
    \centering
    \includegraphics[width=0.47\textwidth]{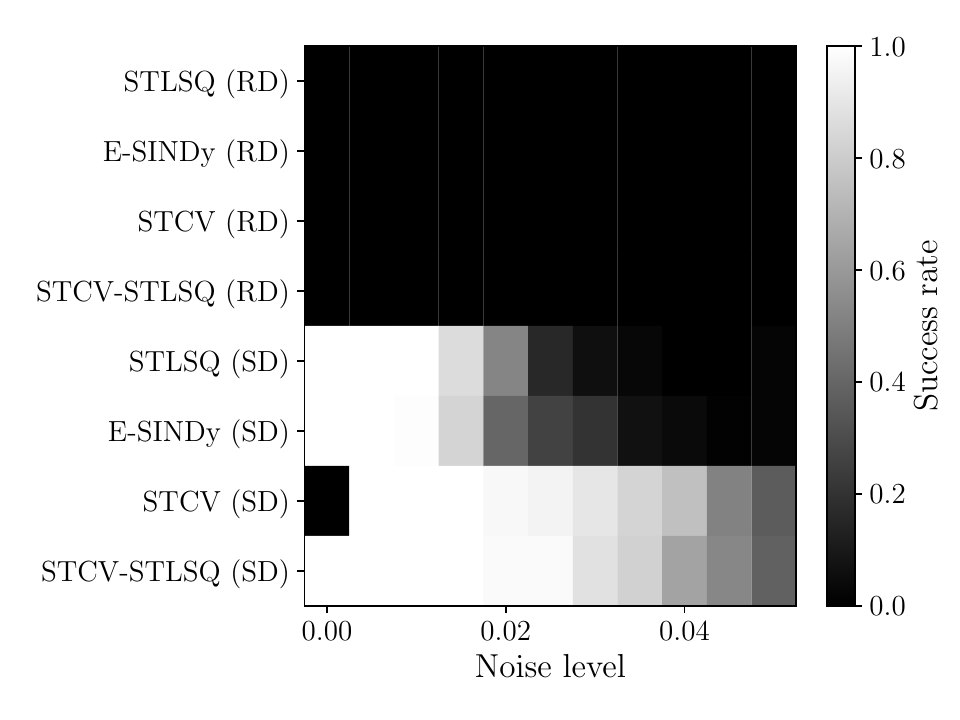}
    \caption{The success rate of various sparse regression algorithms for SINDy on noisy (scaled only, as unscaled cannot be modelled due to numerical limitations) linear oscillator data representing a damaged bearing housing during operation.}
    \label{fig:fig9_bearing}
\end{figure}

The results of the different methods (shown in Figure \ref{fig:fig9_bearing}) demonstrate that STCV and a combined STCV-STLSQ approach significantly outperformed STLSQ and E-SINDy, maintaining a high success rate at noise levels where the other methods were ineffective.

\subsubsection{Half-Car Models}
\begin{figure}[h!]
    \centering
    \begin{subfigure}[b]{0.49\textwidth}
        \centering
        \caption{Linear half-car model}        
        \includegraphics[width=\textwidth]{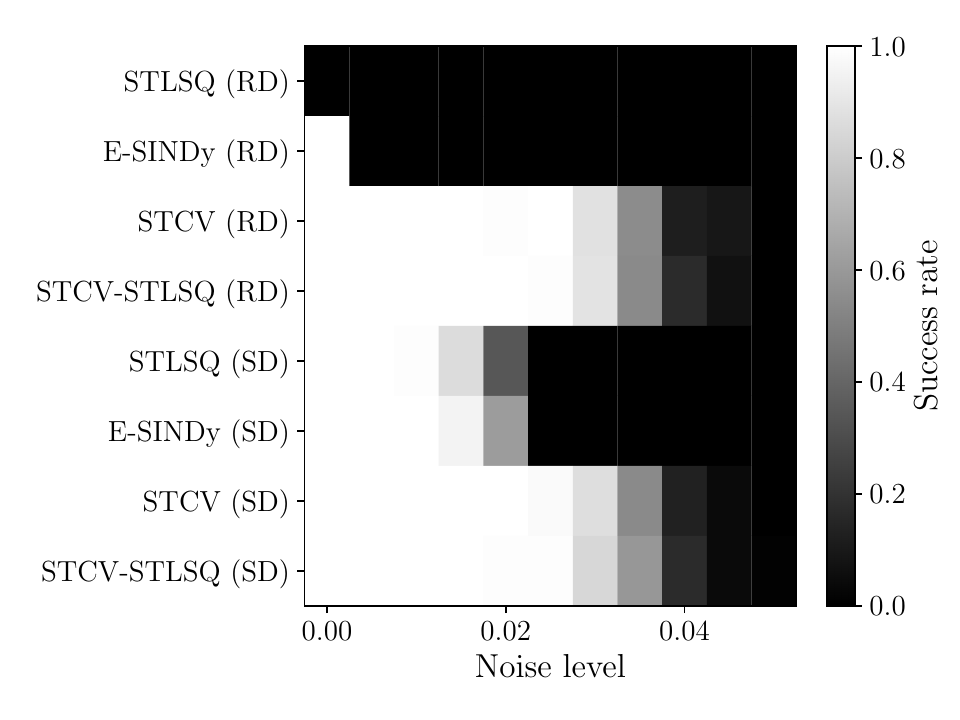} 
        \label{fig:fig10_linear_halfcar}
    \end{subfigure}
    \hfill
    \begin{subfigure}[b]{0.49\textwidth}
        \centering
        \caption{Nonlinear half-car model}        
        \includegraphics[width=\textwidth]{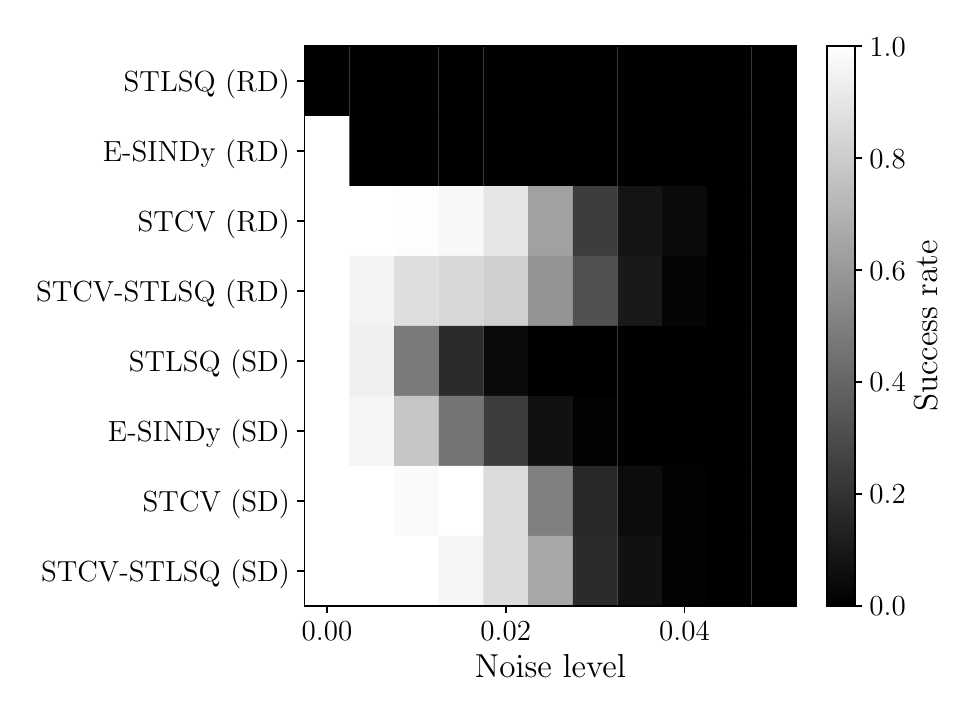}
        \label{fig:fig11_nonlinear_halfcar}
    \end{subfigure}
    \caption{The success rate of various sparse regression algorithms for SINDy on noisy data from the simulated linear and nonlinear half-car models}
\end{figure}

We further tested the algorithms on linear and nonlinear half-car models, which represent higher-dimensional engineering systems. The linear and nonlinear half-car models and their respective parameters are detailed in Appendix ~\ref{app:dataGen}. These eight-state models create a large candidate library, posing a more challenging sparse regression problem. Again, the results in Figures \ref{fig:fig10_linear_halfcar} and \ref{fig:fig11_nonlinear_halfcar} showed that, for both unscaled and normalised data, STCV-based methods consistently provided the most robust performance, correctly identifying the model sparsity far into higher-noise regimes than STLSQ or E-SINDy.

We can see that in all these engineering examples, only STCV or the sequential approach of STCV and STLSQ can correctly identify model sparsity beyond certain noise levels. While E-SINDy improves on STLSQ, the shared point of absolute failure is again observed. Upon further investigation, it was also noted that in many high-noise cases, STCV failed to converge with only one or two additional terms. The lack of improvement when using STCV-STLSQ sequentially shows that the overfit terms are unlikely to regress to small coefficients, even with a fairly small library for STLSQ to sparsify, indicating that the problem is fundamentally difficult to solve at higher noise levels.

More interestingly, we can also observe that for these practical problems and the 2D toy problems with Newtonian-like physics (Van der Pol and Duffing oscillators), the success rate of STCV seems invariant to data scaling (note the shared x-axis scaling for noise levels), except for the bearing system, where normalising was necessary. Additionally, for these Newtonian problems, STCV-based methods consistently outperform E-SINDy and STLSQ.

\subsection{Experimental Validation: Physical Mass-Spring-Damper System}

To complete the investigation, the method was benchmarked using experimental data. This serves as an important demonstration of the necessity of techniques such as the proposed STCV. Unlike the previous benchmarks, only a single noise level could be produced; therefore, the focus is on the model forms recovered.

\begin{figure}[h!]
    \centering
    \includegraphics[width=0.7\textwidth]{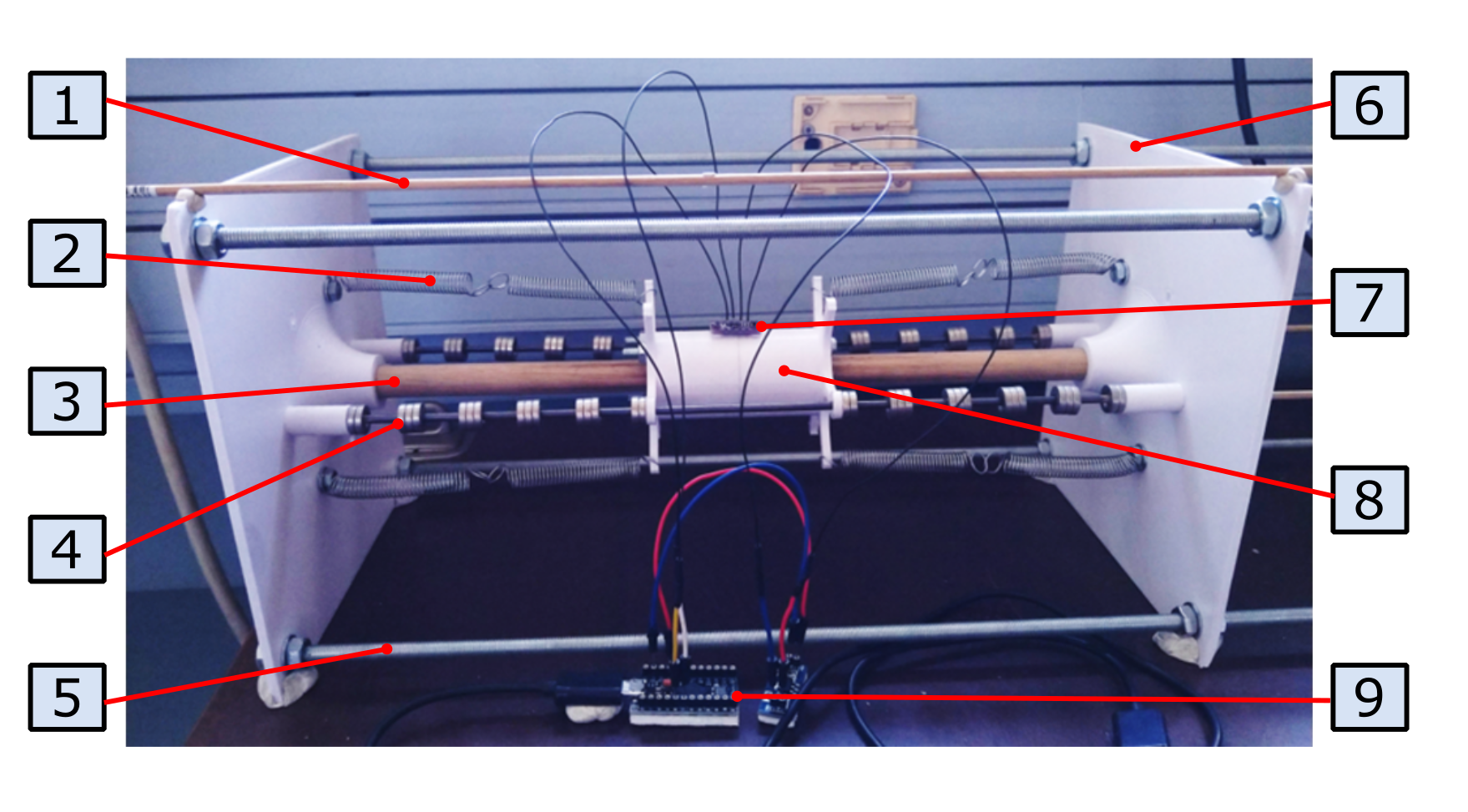}
    \caption{Experimental setup of physical linear/nonlinear mass-spring-damper system. Component list:  1) IMU wiring guide rod, 2) Linear springs, 3) Guide rod for oscillating mass, 4) Nonlinear magnetic springs simulated by stacked magnets on guide rods, 5) Rods securing end plates, 6) End plates holding springs and guide rod, 7) IMU glued to mass, 8) Oscillating mass, 9) Arduino logger.}
    \label{fig:fig12_setup}
\end{figure}

The experimental setup, shown in Figure \ref{fig:fig12_setup}, consists of an oscillating mass. The mass is allowed to oscillate on a linear guide rod with linear bearings, positioned between two end plates. It is connected to the end plates by coiled springs and magnets on both sides. The springs and magnets were arranged such that the system is in static equilibrium when the mass is centred. The setup was designed to have a natural frequency between 1 Hz and 25 Hz, typical of many mechanical machineries such as vehicle suspensions~\cite{ref40}, with the stiffnesses selected to ensure an underdamped response.

In the experimental setup, an LSM6DS3 IMU~\cite{ref39} was used to obtain single-axis acceleration data, sampled at approximately 1.3 kHz using an Arduino Pro Mini (ATmega32u4). The accelerometer was mounted on the oscillating mass and connected with AWG 30 wires, which introduced negligible external dynamics. The LSM6DS3 sensors typically exhibit very low noise levels ($<1\%$). However, tolerances and mechanical play between connected moving components introduce most of the stochastic behaviour observed in the setup.



\begin{figure}[h!]
    \centering
    \includegraphics[width=0.58\textwidth, height=6cm]{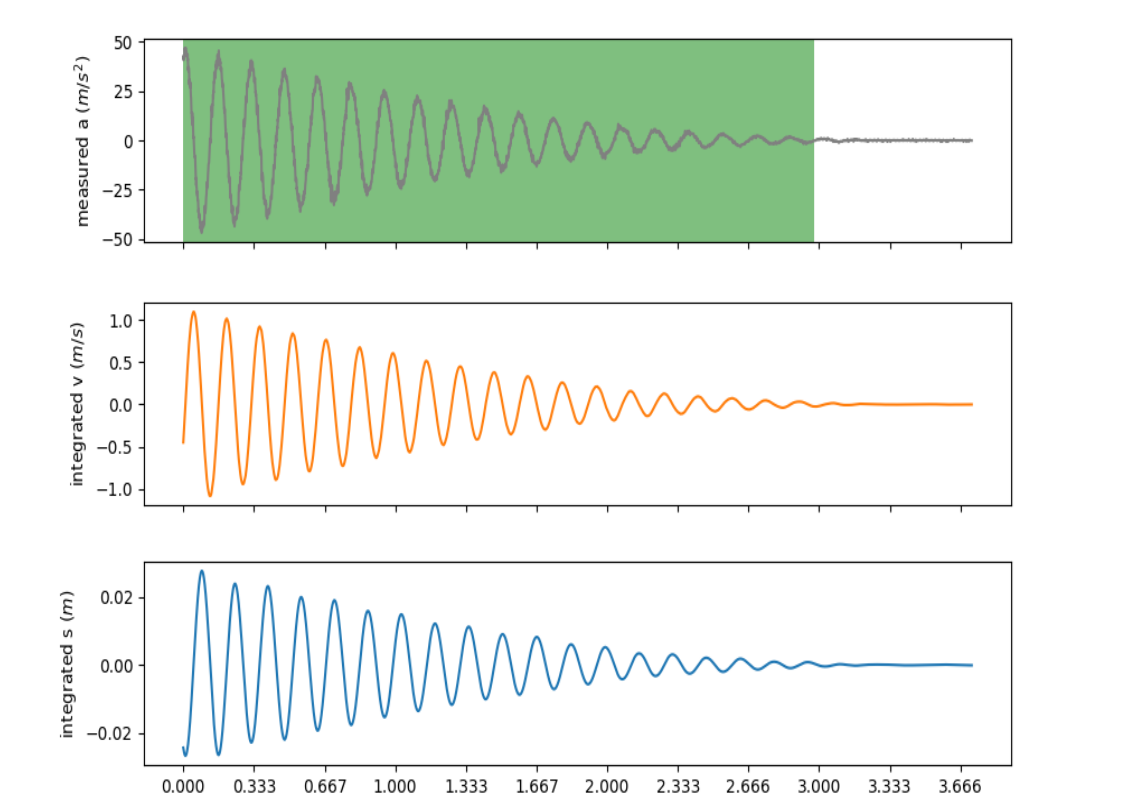}
    \includegraphics[width=0.38\textwidth, height=6cm]{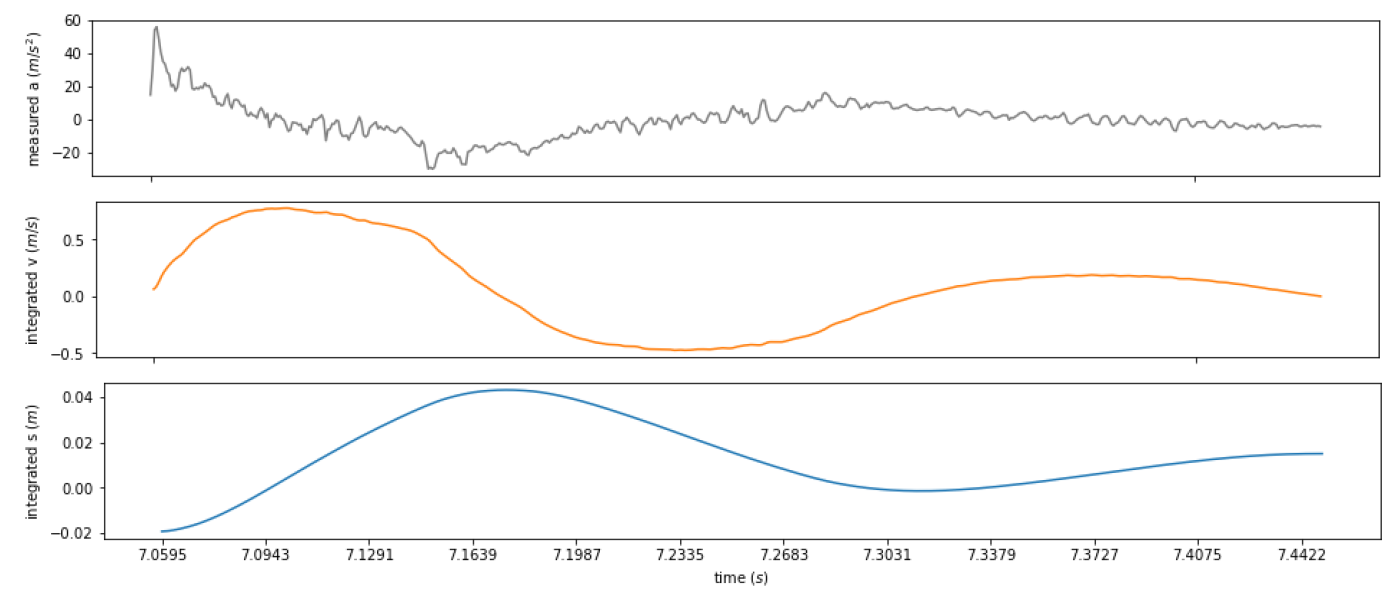}     
    \caption{Measured and integrated data of the physical linear (left) and nonlinear (right) oscillating system. }
    \label{fig:fig13_linear_data_and_nonlinear}
\end{figure}

The measured accelerations, as well as the integrated velocity and displacement responses, are shown in Figure \ref{fig:fig13_linear_data_and_nonlinear} for both the linear and nonlinear systems. The linear system was obtained by removing the magnets, so that only the springs contributed to the system’s stiffness. In contrast, the nonlinear system was obtained by removing the springs and using only the magnets to provide stiffness. The response of the linear system in Figure \ref{fig:fig13_linear_data_and_nonlinear}(left) exhibits the characteristic behaviour of a linear single degree-of-freedom system, while the nonlinear system's response in Figure \ref{fig:fig13_linear_data_and_nonlinear}(right) shows more complex dynamics. The region of data in which the signal-to-noise ratio was deemed high enough for well-conditioned modelling was selected and cropped, as highlighted.

The data are modelled using STLSQ, E-SINDy, and STCV. The hyperparameters used in the algorithms are detailed in Appendix ~\ref{app:modelGen}. The results obtained using the STLSQ, E-SINDy, and STCV methods are summarised in Table \ref{tab:table5_linear_nonlinear_comparison}. The hyperparameters used for STLSQ and E-SINDy to produce the results shown are on the absolute limits of the logical range ($0.9$ and $0.\dot{9}$), and the models cannot be sparsified further without eliminating desired terms. For the linear system, whose model form was determined using an electronic scale for mass, the logarithmic decrement for damping, and a static load test for stiffness, only the STCV method successfully recovered the correct form. In contrast, the STLSQ and E-SINDy algorithms produced models containing dominant spurious terms. For the nonlinear system, the exact model form is unknown. Therefore, the recovered model forms were interrogated. All identified models include nonlinear stiffness terms (e.g., $s^3$) and an increase in damping. However, the STLSQ and E-SINDy models contain more terms than the STCV model, including some that are physically implausible for the system (e.g., $s^2 v$). Hence, it is concluded that the STCV's performance is validated by experimental data as well. 

\begin{table}[h!]
\caption{Comparison of resulting sparse models for the linear and nonlinear experimental data.}
\label{tab:table5_linear_nonlinear_comparison}
\centering
\begin{tabular}{lll}
\toprule
\textbf{Method} & \textbf{Linear system} & \textbf{Nonlinear system} \\
\midrule
Expected &
$\begin{aligned}
    \dot{s} &= +1.0 v \\
    \dot{v} &= -1642 s - 0.143 v
\end{aligned}$ &
Unknown \\
\midrule
STLSQ & 
$\begin{aligned}
    \dot{s} &= +1.0 v \\
    \dot{v} &= -1519.43 s - 2.53 v - 363.36 s^2 + 4.75 sv \\
            &\quad - 218332.31 s^3 - 191.22 s^2v - 228.92 sv^2 \\
            &\quad + 2.99 v^3
\end{aligned}$ &
$\begin{aligned}
    \dot{s} &= +1.0 v \\
    \dot{v} &= -17.541 - 1653.2 s - 20.92 v - 41718.39 s^2 \\
            &\quad + 309.9 sv - 12.02 v^2 - 218332.31 s^3 \\
            &\quad + 12314.87 s^2v + 949.39 sv^2 + 64.41 v^3
\end{aligned}$ \\
\midrule
E-SINDy & 
$\begin{aligned}
    \dot{s} &= +1.0 v \\
    \dot{v} &= -1519.43 s - 2.53 v - 363.36 s^2 \\
            &\quad - 218332.31 s^3 + 191.22 s^2v - 228.92 sv^2 \\
            &\quad + 2.99 v^3
\end{aligned}$ &
$\begin{aligned}
    \dot{s} &= +1.0 v \\
    \dot{v} &= -17.461 - 1567.36 s - 20.69 v - 39480.1 s^2 \\
            &\quad - 9.45 v^2 - 429447.2 s^3 + 5187.76 s^2v \\
            &\quad + 742.63 sv^2 + 50.02 v^3
\end{aligned}$ \\
\midrule
STCV & 
$\begin{aligned}
    \dot{s} &= +1.0 v \\
    \dot{v} &= -1564.26 s - 1.93 v
\end{aligned}$ &
$\begin{aligned}
    \dot{s} &= +1.0 v \\
    \dot{v} &= -17.651 - 1360.26 s - 9.16 v - 33094.72 s^2 \\
            &\quad - 371947.09 s^3
\end{aligned}$ \\
\midrule
STCV\\-STSLQ & 
$\begin{aligned}
    \dot{s} &= +1.0 v \\
    \dot{v} &= -1564.26 s - 1.93 v
\end{aligned}$ &
$\begin{aligned}
    \dot{s} &= +1.0 v \\
    \dot{v} &= -17.651 - 1360.26 s - 9.16 v - 33094.72 s^2 \\
            &\quad - 371947.09 s^3
\end{aligned}$ \\
\bottomrule
\end{tabular}
\end{table}

\begin{figure}[h!]
    \centering
    \includegraphics[width=0.84\textwidth]{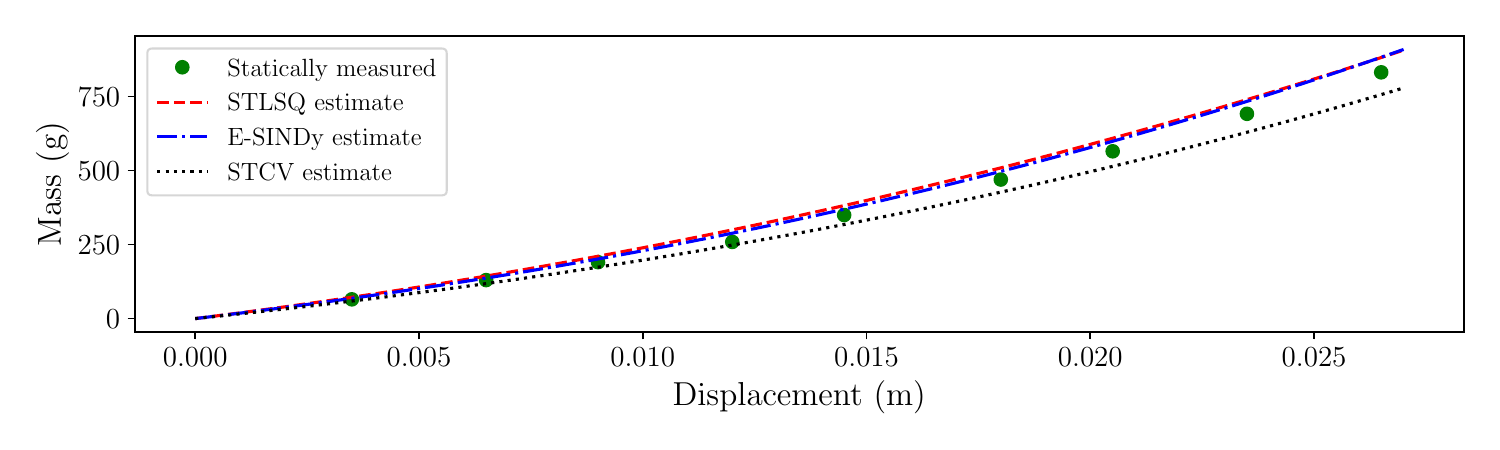}
    \caption{Stiffness of nonlinear setup, statically measured vs. dynamically estimated by STCV, STLSQ, and E-SINDy.}
    \label{fig:fig15_stiffness}
\end{figure}

To verify the fidelity of the answers, the statically measured stiffness is compared against the nonlinear returning force. The cubic function returned by STCV, as well as the STSLQ and E-SINDy estimates obtained by setting $v = 0$, are used in this comparison, with the results shown in Figure \ref{fig:fig15_stiffness}. The STLSQ and E-SINDy methods predict a stiffer system than the measured one, while the STCV method yields a lower stiffness. Since the final regression scheme for the model coefficients is identical for all methods, this difference arises for the model form identified. It is possible that static measurements of stiffness overestimate the stiffness of such a system due to static friction.

\subsection{Analysis of Model Selection Characteristics}

To better understand the behaviour of STCV, we conducted a bias test on a system with competing linear ($k_1 s$) and cubic ($k_2 s^3$) stiffness terms using the following nonlinear oscillator 
\begin{equation} \label{eq:study}
\begin{aligned}
    \dot{s} &= v \\
    \dot{v} &= -10 v - k_1 s - k_2 s^3
\end{aligned}
\end{equation}
where $k_1$ and $k_2$ are the linear and cubic stiffness coefficients. We set $k_1=300$ and $k_2=1000$ to yield similar natural frequencies, thereby isolating selection bias from sampling-rate effects.

\begin{figure}[h!]
    \centering
    \includegraphics[width=0.9\textwidth]{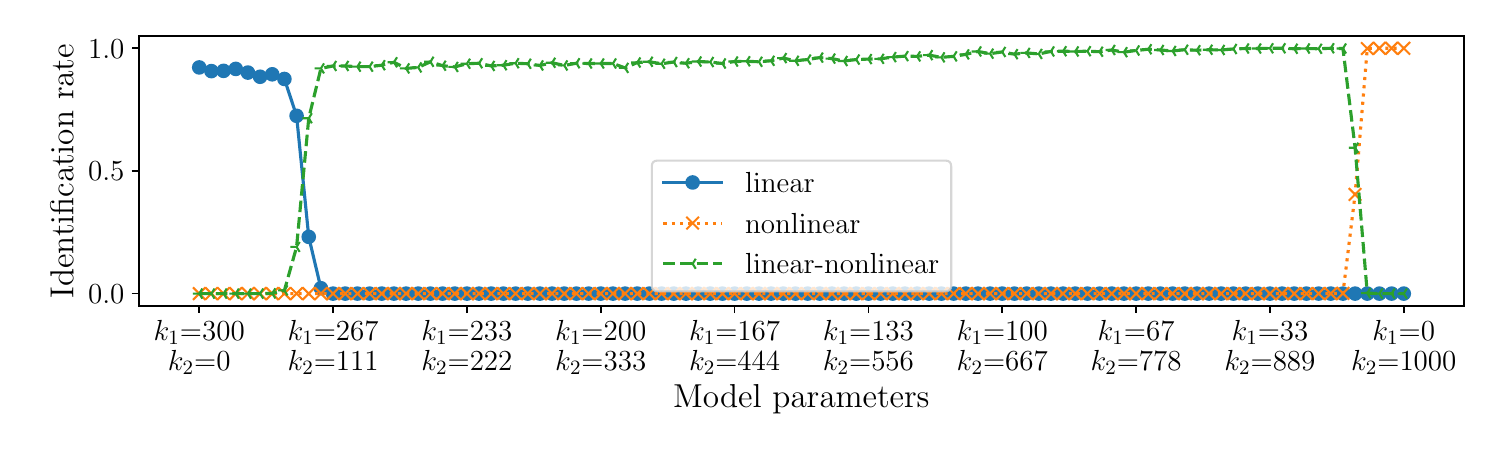} 
    \caption{STCV nonlinearity bias test. 100 different models based on Equation \eqref{eq:study}, with varying values of $k_1$ and $k_2$, are produced. The values of $k_1$ and $k_2$ are ramped inversely, so the underlying model transitions from being strongly linear to strongly nonlinear. For each model, 1000 noisy datasets are produced and modelled using STCV. The model identification rate is the success rate at which STCV fits either a linear model (only $k_1$ present), a linear-nonlinear model (both $k_1$ and $k_2$ present), or a nonlinear model (only $k_2$ present). $k_1$ and $k_2$ are tuned to have similar induced natural frequencies, and this experiment should help determine whether STCV is biased toward fitting linear or nonlinear models.}
    \label{fig:fig16_bias}
\end{figure}

We test 100 combinations, ramping $k_1$ from 300 down to 0 while concurrently ramping $k_2$ from 0 up to 1000. For each combination, 1000 noisy datasets (1\% noise, refer to Appendix ~\ref{app:dataGen} for data generation procedure) are generated, normalised, and modelled using STCV with a CP threshold of 0.3. Figure \ref{fig:fig16_bias} plots the identification rate for linear-only, nonlinear-only, and mixed linear-nonlinear models across this parameter sweep.

The results in Figure \ref{fig:fig16_bias} show STCV is effective at finding the correct model form, though it misclassifies some edge cases where coefficients are very small. This is a consequence of the STCV mechanism, which requires a minimum term strength (coefficient magnitude) for detection.

\begin{figure}[h!]
    \centering
    \includegraphics[width=0.9\textwidth]{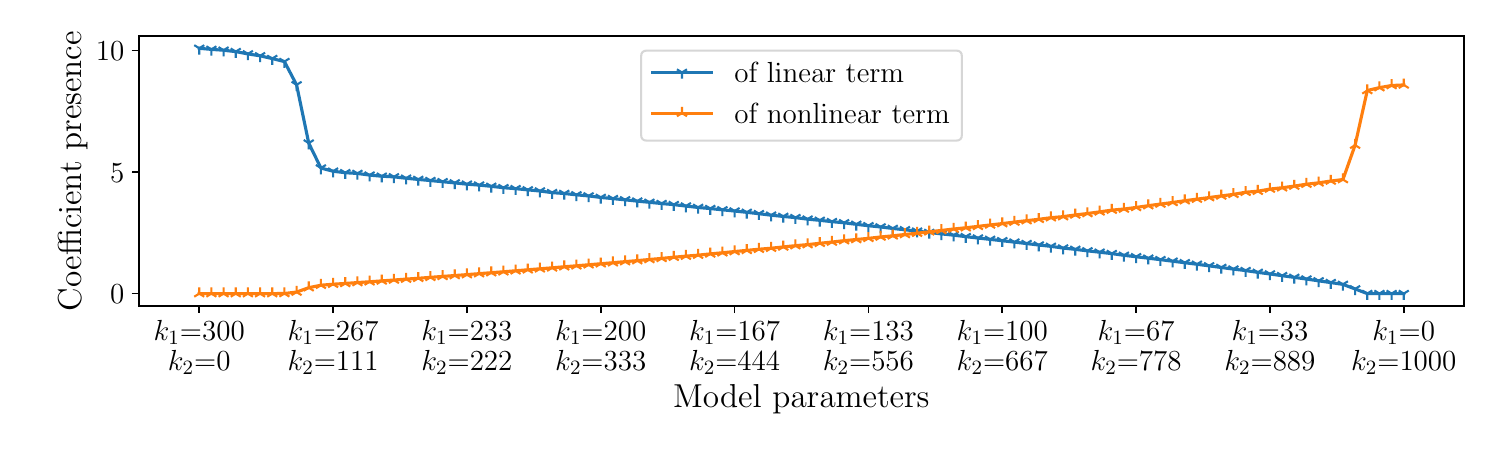} 
    \caption{Average coefficient presence (CP) values of the linear and nonlinear terms from the STCV bias test.}
    \label{fig:fig17_cp}
\end{figure}

Figure \ref{fig:fig17_cp} shows the average Coefficient Presence (CP) values. The CP, although dimensionless, is clearly dependent on the magnitude of the coefficient and its contribution to the dynamics. The method exhibits a desirable "snapping" behaviour as would be expected of an $L_{0}$-based method. As one term's strength increases, its CP value rises sharply, while the CP value of the weaker term drops rapidly to zero. This allows the model to snap cleanly between forms rather than transitioning smoothly. The CP plots do not intersect perfectly at the midpoint, which is attributed to the imperfect scaling of the $k_1$ and $k_2$ effects, rather than to an inherent bias. The results show no conclusive evidence of STCV bias towards or against nonlinear terms.


\section{Discussion and Future Outlook}
\label{sec:DiscussionFutureOutlook}
The benchmarks demonstrate that STCV successfully addresses a critical limitation of SINDy. Replacing magnitude-based thresholding with a statistical validity check provides robust model identification from normalised data, a common necessity in practice. STCV is therefore a crucial component in the robust data-driven discovery ecosystem.

A key area for future work is combining STCV with other robust methods. One option could be the natural extension of SR3 from STLSQ, where the algorithm is redefined as an objective function that can be minimised, making the method modular. STCV and weak-form SINDy (WSINDy) are not mutually exclusive, as they address orthogonal failure modes: WSINDy robustifies derivative estimation, while STCV robustifies the regression against scaling issues. A combined WSINDy-STCV pipeline could offer unparalleled robustness.

Compared to Bayesian SINDy frameworks~\cite{ref13}~\cite{ref14}, STCV offers a significant computational advantage. Its use of an efficient, closed-form BLR calculation avoids expensive MCMC sampling, making it suitable for large-scale applications.

Current limitations include underperformance on perfectly noise-free data (mitigated by injecting small amounts of noise) and the need for manual hyperparameter selection ($\lambda_{CP}, \gamma$). Future work should focus on automated, data-driven methods for hyperparameter tuning, perhaps adapting techniques from ADAM-SINDy~\cite{ref21}.

Further research could also integrate STCV with frameworks such as constrained SINDy~\cite{ref24} or extend its statistical foundation to improve uncertainty quantification (UQ). Combining STCV's robust selection with the rigorous, distribution-free guarantees of Conformal Prediction~\cite{ref26} is a promising direction for safety-critical applications.

\section{Conclusion}
\label{sec:Conclusion}
The sensitivity of sparse identification methods to routine data preprocessing steps, such as normalisation, poses a significant barrier to their reliable and automated application. This work has demonstrated that the interaction between data scaling and measurement noise can systematically distort the coefficient landscape of a dynamical system, causing magnitude-based regression algorithms like STLSQ to fail by selecting dense, physically incorrect models.

We have introduced and validated the Sequential Thresholding of Coefficient of Variation (STCV) algorithm as a direct and effective solution to this problem. By employing a dimensionless, statistically grounded metric, the Coefficient Presence-STCV approach selects model terms based on their consistency and statistical significance rather than their magnitude. This makes the discovery process largely invariant to data scaling. Through extensive testing across numerical and physical systems, we have shown that STCV consistently outperforms standard SINDy approaches on normalised, noisy data, successfully identifying parsimonious, interpretable models in regimes where other methods fail, whilst being computationally efficient. The successful identification of the correct governing equation from a real-world physical experiment underscores its practical utility. By providing a magnitude-free regression framework, STCV makes SINDy a more robust, reliable, and practical tool for discovering knowledge from real-world scientific and engineering data, thereby advancing the overarching goal of automated scientific discovery. Being a general-purpose sparse regression algorithm, it also contributes to the general field of regression.

\appendix
\section{Code Repository}
\label{app:Code}
The code and models used to generate the data in this paper are available here: \url{https://github.com/RautJ/STCV}.

\section{Experimental Data Generation}
\label{app:dataGen}

Multiple numerical experiments were conducted to demonstrate the performance of various regression schemes for SINDy. Large datasets were generated for these experiments. This appendix provides additional details on the data generation procedure for each experiment.

Firstly, noise-free data are generated for the dynamical systems by numerically integrating their governing ODEs as an Initial Value Problem using SciPy's solve\_ivp function. The bearing and half-car models are special, as instead of an unsteady initial condition, an initial excitation is applied. Only post-excitation data is measured. Refer to the system discussions for details of these initial conditions. The ODEs, alongside integrator setup are listed in Table ~\ref{tab:tableB1}.

\begin{table}[h!]
\caption{Numerical integration parameters for dynamical systems.}
\label{tab:tableB1}
\centering
\begin{tabular}{lllllll}
\toprule
\textbf{System} & \textbf{ODE} & \textbf{Integrator} & \textbf{Initial value} & \textbf{Sampling rate} & \textbf{Period}  & \textbf{Tolerance} \\
\midrule
Lorenz & 
$\begin{aligned}
    Ref.~\cite{ref41} \\
\end{aligned}$ &
$\begin{aligned}
    RK45 \\
\end{aligned}$ &
$\begin{aligned}
    [0,1,20] \\
\end{aligned}$ &
$\begin{aligned}
    100Hz \\
\end{aligned}$ &
$\begin{aligned}
    10s \\
\end{aligned}$ &
$\begin{aligned}
    10^{-12} \\
\end{aligned}$ \\
\midrule
R\"{o}ssler & 
$\begin{aligned}
    Ref.~\cite{ref41} \\
\end{aligned}$ &
$\begin{aligned}
    RK45 \\
\end{aligned}$ &
$\begin{aligned}
    [14,8,0] \\
\end{aligned}$ &
$\begin{aligned}
    1kHz \\
\end{aligned}$ &
$\begin{aligned}
    10s \\
\end{aligned}$ &
$\begin{aligned}
    10^{-12} \\
\end{aligned}$ \\
\midrule
van der Pol & 
$\begin{aligned}
    Ref.~\cite{ref41} \\
\end{aligned}$ &
$\begin{aligned}
    LSODA \\
\end{aligned}$ &
$\begin{aligned}
    [2,0] \\
\end{aligned}$ &
$\begin{aligned}
    100Hz \\
\end{aligned}$ &
$\begin{aligned}
    30s \\
\end{aligned}$ &
$\begin{aligned}
    default \\
\end{aligned}$ \\
\midrule
Duffing & 
$\begin{aligned}
    Ref.~\cite{ref41} \\
\end{aligned}$ &
$\begin{aligned}
    LSODA \\
\end{aligned}$ &
$\begin{aligned}
    [1,0] \\
\end{aligned}$ &
$\begin{aligned}
    100Hz \\
\end{aligned}$ &
$\begin{aligned}
    10s \\
\end{aligned}$ &
$\begin{aligned}
    default \\
\end{aligned}$ \\
\midrule
Bearing & 
$\begin{aligned}
    Equation ~\ref{eqn:bearing} \\
\end{aligned}$ &
$\begin{aligned}
    RK45 \\
\end{aligned}$ &
$\begin{aligned}
    [0,0] \\
\end{aligned}$ &
$\begin{aligned}
    1MHz \\
\end{aligned}$ &
$\begin{aligned}
    5\mu s \\
\end{aligned}$ &
$\begin{aligned}
    10^{-12} \\
\end{aligned}$ \\
\midrule
Linear HC & 
$\begin{aligned}
    Ref. ~\cite{ref42} \\
\end{aligned}$ &
$\begin{aligned}
    Radau \\
\end{aligned}$ &
$\begin{aligned}
    [0,...,0] \\
\end{aligned}$ &
$\begin{aligned}
    5kHz \\
\end{aligned}$ &
$\begin{aligned}
    2.12s \\
\end{aligned}$ &
$\begin{aligned}
    10^{-12} \\
\end{aligned}$ \\
\midrule
Non-Linear HC & 
$\begin{aligned}
    Ref. ~\cite{ref43} \\
\end{aligned}$ &
$\begin{aligned}
    Radau \\
\end{aligned}$ &
$\begin{aligned}
    [0,...,0] \\
\end{aligned}$ &
$\begin{aligned}
    5kHz \\
\end{aligned}$ &
$\begin{aligned}
    2.12s \\
\end{aligned}$ &
$\begin{aligned}
    10^{-12} \\
\end{aligned}$ \\
\midrule
Oscillator & 
$\begin{aligned}
    Equation~\ref{eq:study} \\
\end{aligned}$ &
$\begin{aligned}
    RK45 \\
\end{aligned}$ &
$\begin{aligned}
    [0.15,9.89] \\
\end{aligned}$ &
$\begin{aligned}
    100Hz \\
\end{aligned}$ &
$\begin{aligned}
    5s \\
\end{aligned}$ &
$\begin{aligned}
    10^{-12} \\
\end{aligned}$ \\
\bottomrule
\end{tabular}
\end{table}

Upon generating time signals for each problem's state variables, we apply measurement noise. For each problem, the time signals for each variable are normalised (i.e., scaled so that the maximum absolute value is unity). On these normalised signals, Gaussian white noise is added. The intent is to simulate a percentage noise-floor. To simulate this, a 1\% noise-floor is considered to be $N(0,0.005)$ (i.e., the 95th percentile is set to be a value of 1\%. This noisy, normalised signal is scaled back to the raw scales to produce a noisy, raw signal. This process is repeated 1000 times, with unique noise realisations at each noise level.

To conduct SINDy, Equation ~\ref{eq:one}, gradient data is required as well. This is numerically inferred from the noisy time signals through second-order numerical differentiation. Additionally, a set of basis functions must be selected, and the trajectory data must be passed through them to generate the design matrix. In this study, a polynomial pool of order 3 is used for all experiments.

Special Note: Half-car models are treated uniquely. Since the governing ODE is presented in a factorised form, we preserve this form in the SINDy-returned model by measuring the relative displacements and velocities between the system's lumped masses rather than the direct displacements and velocities of the lumped masses.

\section{Regressor Setups}
\label{app:modelGen}

Multiple numerical experiments were conducted to demonstrate the performance of various regression schemes for SINDy. Appendix~\ref{app:dataGen} details the generation of SINDy-ready data for these experiments. This appendix details how the various regressors (STLSQ, E-SIND, STCV) were set up and the hyperparameters used in the benchmarking experiments.

The STLSQ implementation used in this study is identical to that in PySINDy~\cite{ref12}, except for 2 additional steps. These steps are the following. Once the linear problem is defined, we solve the normal equations of the linear system instead of solving the system directly. Secondly, we use a direct linear solver from NumPy instead of an adaptive solver. These modifications have been noted to insignificantly alter the performance of STLSQ as compared to the implementation in PySINDy.

E-SINDy has many flavours, and in this study, we explore only the data-bagging variant with fixed subsampling parameters (100 subsamples, 63.2\% overlap). The intent is to provide a benchmark that may be used to compare the results of E-SINDy presented in this paper to those in the original E-SINDy study.

The STCV implementation is as described in Algorithm ~\ref{tab:stcv_alg}.

The benchmarking experiment compares the performance of the various methods under increasing noise and explores their potential to recover the correct underlying model form from data. This is done by performing a grid search on the sparsifying hyperparameter and checking whether the correct model form was ever recovered.

STLSQ offers 1 main sparsifying hyperparameter - the threshold value, $\lambda$. All methods use STLSQ internally. E-SINDy further adds a sparsifying hyperparameter - the inclusion probability threshold. Our STCV method uses the CP threshold as the primary sparsifying hyperparameter. All methods have an optional ridge penalty hyperparameter; however, this is set to a fixed value of $10^{-16}$. 10 internal ramping steps are used in the STCV method for the benchmark. For the additional STCV-STLSQ method, the ridge penalty (both final and initial; see Algorithm~\ref{tab:stcv_alg}) is raised to a large power.

The ranges of the sparsifying hyperparameters of STLSQ and E-SINDy are logically bound. For STLSQ, we perform a grid search over linearly spaced threshold values from 1\% to 90\% of the smallest model coefficient value; for the Lorenz problem, this ranges from 0.01 to 0.9 since the smallest model coefficient is 1. For E-SINDy, we scan for an inclusion threshold from 10\% to 90\%. The STLSQ threshold to be defined within E-SINDy is set to the value that achieved the highest modelling success rate. The intent is to provide each method with the best possible chance of success while limiting the dimensionality of the grid search. When STLSQ yields no successes, choosing the optimal threshold is not defined but is inconsequential, as E-SINDy has no successes for these problems regardless of the STLSQ threshold value.

With STCV, the CP threshold does not have a logical, bounded range. We manually identify a range for each problem through trial and error, which yields good results. The ranges used in the benchmarking analysis for each problem are listed in Table ~\ref{tab:tableC1}. Note that the CP ranges are scanned with geometric spacing, yet with 10 steps as with the other algorithms. When using STLSQ within STCV, we set the STLSQ threshold to the lowest value in the range used by the STLSQ grid search (e.g., for the Lorenz system, this is 0.01).

Additionally, in Table ~\ref{tab:tableC1}, we provide the ridge penalty used in the STCV-STLSQ method pursued. In this method, for each step in the CP threshold scan, the STLSQ method is run internally as a nested scan. The ridge penalty is applied only to the STCV method when STLSQ is used internally.

\begin{table}[h!]
\caption{STCV hyperparameters used for the benchmarking experiment.}
\label{tab:tableC1}
\centering
\begin{tabular}{lll}
\toprule
\textbf{System} & \textbf{CP threshold range} & \textbf{Ridge penalty (STCV-STLSQ)} \\
\midrule
Lorenz & 
$\begin{aligned}
    [0.001,1] \\
\end{aligned}$ &
$\begin{aligned}
    10^{-1} \\
\end{aligned}$ \\
\midrule
Lorenz (normalised) & 
$\begin{aligned}
    [0.001,3] \\
\end{aligned}$ &
$\begin{aligned}
    10^{-1} \\
\end{aligned}$ \\
\midrule
R\"{o}ssler & 
$\begin{aligned}
    [0.001,0.8] \\
\end{aligned}$ &
$\begin{aligned}
    10^{-1} \\
\end{aligned}$ \\
\midrule
R\"{o}ssler (normalised) & 
$\begin{aligned}
    [0.1,1.2] \\
\end{aligned}$ &
$\begin{aligned}
    10^{-5} \\
\end{aligned}$ \\
\midrule
van der Pol & 
$\begin{aligned}
    [0.001,0.2] \\
\end{aligned}$ &
$\begin{aligned}
    10^{-1} \\
\end{aligned}$ \\
\midrule
van der Pol (normalised) & 
$\begin{aligned}
    [0.001,0.2] \\
\end{aligned}$ &
$\begin{aligned}
    10^{-1} \\
\end{aligned}$ \\
\midrule
Duffing & 
$\begin{aligned}
    [0.001,0.2] \\
\end{aligned}$ &
$\begin{aligned}
    10^{2} \\
\end{aligned}$ \\
\midrule
Duffing (normalised) & 
$\begin{aligned}
    [0.001,0.2] \\
\end{aligned}$ &
$\begin{aligned}
    10^{2} \\
\end{aligned}$ \\
\midrule
Bearing (normalised)& 
$\begin{aligned}
    [0.001,0.03] \\
\end{aligned}$ &
$\begin{aligned}
    10^0 \\
\end{aligned}$ \\
\midrule
Linear HC & 
$\begin{aligned}
    [0.11,0.14] \\
\end{aligned}$ &
$\begin{aligned}
    10^{-16} \\
\end{aligned}$ \\
\midrule
Linear HC (normalised) & 
$\begin{aligned}
    [0.11,0.14] \\
\end{aligned}$ &
$\begin{aligned}
    10^{-16} \\
\end{aligned}$ \\
\midrule
Non-Linear HC & 
$\begin{aligned}
    [0.03,0.05] \\
\end{aligned}$ &
$\begin{aligned}
    10^{-4} \\
\end{aligned}$ \\
\bottomrule
Non-Linear HC (normalised) & 
$\begin{aligned}
    [0.03,0.05] \\
\end{aligned}$ &
$\begin{aligned}
    10^{-4} \\
\end{aligned}$ \\
\bottomrule
\end{tabular}
\end{table}

A physical experiment is conducted to analyse the potential real-world performance of the three algorithms. The values of the modelling hyperparameters used to produce the results are shown in Table ~\ref{tab:table5_linear_nonlinear_hyperparameters}. Take special note of the STLSQ and E-SINDy thresholding, as they are set to a value on the edge of their logical ranges.

\begin{table}[h!]
\caption{Comparison of modelling hyperparameters used for the linear and nonlinear experimental data.}
\label{tab:table5_linear_nonlinear_hyperparameters}
\centering
\begin{tabular}{lll}
\toprule
\textbf{Method} & \textbf{Linear system} & \textbf{Nonlinear system} \\
\midrule
STLSQ & 
$\begin{aligned}
    \gamma &= 10^{-16} \\
    \lambda &= 0.9 \\
\end{aligned}$ &
$\begin{aligned}
    \gamma &= 10^{-16} \\
    \lambda &= 0.9 \\
\end{aligned}$ \\
\midrule
E-SINDy & 
$\begin{aligned}
    \gamma &= 10^{-16} \\
    \lambda &= 0.9 \\
    Incl. thres. &= (1-10^{-16})
\end{aligned}$ &
$\begin{aligned}
    \gamma &= 10^{-16} \\
    \lambda &= 0.9 \\
    Incl. thres. &= (1-10^{-16})
\end{aligned}$ \\
\midrule
STCV & 
$\begin{aligned}
    \gamma_0 &= 10^{-16} \\
    \gamma_f &= 10^{-16} \\
    \lambda &= 0.01 \\
    CP &= 0.3 \\
\end{aligned}$ &
$\begin{aligned}
    \gamma_0 &= 10^{-16} \\
    \gamma_f &= 10^{-16} \\
    \lambda &= 0.01 \\
    CP &= 0.4 \\
\end{aligned}$ \\
\bottomrule
\end{tabular}
\end{table}

\section*{Declaration of generative AI and AI-assisted technologies in the manuscript preparation process}
During the preparation of this work, the author(s) used generative artificial intelligence tools for language editing and for assisting with phrasing and clarity of presentation. After using this tool/service, the author(s) reviewed and edited the content as needed and take(s) full responsibility for the content of the published article.


\end{document}